\documentclass[twoside]{article} \usepackage{aistats2017}

\usepackage[utf8]{inputenc} % allow utf-8 input
\usepackage[T1]{fontenc}    % use 8-bit T1 fonts
\usepackage{hyperref}       % hyperlinks
\usepackage{url}            % simple URL typesetting
\usepackage{booktabs}       % professional-quality tables
\usepackage{amsfonts}       % blackboard math symbols
\usepackage{nicefrac}       % compact symbols for 1/2, etc.
\usepackage{microtype}      % microtypography

\usepackage{graphicx}
\usepackage{times}

\usepackage{amssymb,amsmath,epsfig,bm,comment,url}
\usepackage[ruled,vlined]{algorithm2e}
\usepackage{bbding}
\usepackage{booktabs}
\usepackage{nicefrac}

\newcommand{\defeq}{\stackrel{\mathrm{def}}{=}}

\newcommand{\E}[2]{\mathrm{E}_{#1}\left[#2\right]}
\renewcommand{\vec}[1]{\bm{#1}}

\newcommand{\mat}[1]{\bm{\mathrm{#1}}}

\newcommand{\nrange}[2]{\{#1,..,#2\}}

\begin{document}

\twocolumn[

\aistatstitle{Compacting Neural Network Classifiers via Dropout Training}

%\aistatsauthor{ Anonymous Author 1 \And Anonymous Author 2 \And Anonymous Author 3 }
\aistatsauthor{Yotaro Kubo \And George Tucker \And Simon Wiesler}

\aistatsaddress{ Alexa Machine Learning, Amazon } 
%\aistatsaddress{ Unknown Institution 1 \And Unknown Institution 2 \And Unknown Institution 3 }
]

\begin{abstract}
We introduce \emph{dropout compaction}, a novel method for training feed-forward neural networks which realizes the performance gains of training a large model with dropout regularization, yet extracts a compact neural network for run-time efficiency. In the proposed method, we introduce a sparsity-inducing prior on the per unit dropout retention probability so that the optimizer can effectively prune hidden units during training. By changing the prior hyperparameters, we can control the size of the resulting network. We performed a systematic comparison of dropout compaction and competing methods on several real-world speech recognition tasks and found that dropout compaction achieved comparable accuracy with fewer than 50\% of the hidden units, translating to a 2.5x speedup in run-time.
\end{abstract}

\section{Introduction}

Dropout~\cite{hinton2012improving,srivastava2014dropout} is a well-known regularization method that has been used very successfully for feed-forward neural networks. Training large models with strong regularization from dropout provides state-of-the-art performance on numerous tasks (e.g., \cite{krizhevsky2012imagenet,dahl2013improving}). The method inserts a ``dropout'' layer which stochastically zeroes individual activations in the previous layer with probability $1-p$ during training. At test time, these stochastic layers are deterministically approximated by rescaling the output of the previous layer by $p$ to account for the stochastic dropout during training. Empirically, dropout is most effective when applied to large models that would otherwise overfit~\cite{srivastava2014dropout}. %, because the empirical loss degrades by introducing stochastic noise to hidden layer outputs.
%Using dropout usually requires the use of a large model \cite{srivastava2014dropout} because it degrades the empirical loss by introducing stochastic noise to hidden layer outputs.
Although training large models is usually not an issue anymore with multi-GPU training (e.g.,~\cite{seide2014sgd,strom2015scalable}), model size must be restricted in many applications to ensure efficient test time evaluation.

In this paper, we propose a novel form of dropout training that provides the performance benefits of dropout training on a large model while producing a compact model for deployment. Our method is inspired by annealed dropout 
training~\cite{rennie2014annealed}, where the retention probability $p$ is slowly annealed to $1$ over multiple epochs.
For our method, we introduce an independent retention probability parameter for each hidden unit with a bimodal prior distribution sharply peaked at $0$ and $1$ to encourage the posterior retention probability to converge to either $0$ or $1$. Similarly to annealed dropout, some units will converge to never being dropped, however, unlike annealed dropout, some units will converge to always being dropped. These units can be removed from the network without accuracy degradation. The annealing schedule and compaction rate of the resulting network can be controlled by changing the hyperparameters of the prior distribution.

In general, model compaction has been well investigated. Conventionally, this problem is addressed by sparsifying the weight matrices of the neural network. For example,~\cite{lecun1989optimal} introduced a pruning criterion based on the second-order derivative of the objective function. L1 regularization is also widely used to obtain sparse weight matrices. However, these methods only achieve weight-level sparsity. Due to the relative efficiency of dense matrix multiplication compared to sparse matrix multiplication, these approaches do not improve test time efficiency without degrading accuracy. In contrast, our proposed method directly reduces the dimension of the weight matrices.

In another approach, singular value decomposition (SVD) is used to obtain approximate low-rank representations of the weight matrices~\cite{xue2013restructuring}.
SVD can directly reduce the dimensionality of internal representations, however, it is typically implemented as a linear bottleneck layer, which introduces additional parameters. This inefficiency requires additional compression, which degrades performance. For example, if we are to compress the number of parameters by half, the SVD compacted model would have to restrict the internal dimensionality by 25\%.
Knowledge distillation can also be used to transfer knowledge from larger models to a small model~\cite{hinton2015distilling}.
However, this requires two separate optimization steps that makes it difficult to directly apply to existing optimization configurations.
\cite{murray2015auto} introduced additional multiplicative parameters to the output of the hidden layers and regularize these parameters with an L1 penalty. This method is conceptually similar to dropout compaction in that both methods introduce regularization to reduce the dimensionality of internal representations. Dropout compaction can be seen as an extension of this method by replacing L1 regularization with dropout-based regularization.

Similar to the SVD-based compaction technique, there are several approaches for achieving faster evaluation of neural networks by assuming a certain structure in the weight matrices.
For example, \cite{vikas2016structured} introduces Toeplitz weight matrices as a building block of neural networks.
\cite{moczulski2016ACDC} defines a structured matrix via discrete cosine transform (DCT) for enabling fast matrix multiplication via a fast Fourier transform algorithm.
These methods successfully reduce the computational cost for neural network prediction; however, since those methods restrict the parameter space prior to training, these method may restrict the flexibility of neural networks.
Our proposed method attempts to jointly optimize the structure of the neural network and its parameters. This way, the optimization algorithm can determine an effective low-dimensional representation of the hidden layers.

Several approaches for reducing numerical precision to speed up training and evaluation have been proposed~\cite{gupta2015deep,courbariaux2014low,courbariaux2015binaryconnect}.
Most of these approaches are complemental with our method since our proposed method only changes the dimensionality of hidden layers, but the structure of the network stays the same. In particular, we also apply a basic weight quantization technique in our experiments on automatic speech recognition tasks.

The remainder of this paper is as follows: In Section \ref{Sec:Dropout}, we introduce a probabilistic formulation of dropout training and cast it as an ensemble learning method. Then, we derive a method for optimizing the dropout retention probabilities $p$ in Section \ref{Sec:ProbOpt}. In Section \ref{Sec:Experiments}, we present experimental results. In Section \ref{Sec:Conclusion}, we conclude the paper and suggest future extensions.

\section{Conventional Dropout Training} \label{Sec:Dropout}

In this section, we describe conventional dropout training for feed-forward neural networks. 
Hereafter, we denote the input vectors by $\mathcal{X} = \{\vec{x}_1, \vec{x}_2, \cdots \vec{x}_T\}$, the target labels by $\mathcal{K} = \{k_1, k_2, \cdots k_T\}$, and the parameters of a neural network by $\Theta \defeq \{\mat{W}^{(\ell)} \in \mathbb{R}^{D^{(\ell)} \times D^{(\ell-1)}}, \vec{b}^{(\ell)} \in \mathbb{R}^{D^{(\ell)}} | \ell \in \nrange{1}{L}\}$ where $L$ is the number of layers and $D^{(\ell)}$ is the number of output units in the $\ell$-th layer.

\subsection{Training}

Dropout training can be viewed as the optimization of a lower bound on the log-likelihood of a probabilistic model.

By introducing a set of random mask vectors $\mat{M} \defeq \{\vec{m}^{(\ell)} \in \{0, 1\}^{D^{(\ell)}} | \ell \in \nrange{0}{L}\}$ which defines a subset of hidden units to be zeroed, the output of a dropout neural network can be expressed as follows:
\begin{equation}
  \begin{split}
    \vec{h}^{(0)}(\vec{x}_t; \mat{M}) &= \vec{m}^{(0)} \odot \vec{x}_t \;, \\
    \vec{h}^{(\ell)}(\vec{x}_t; \mat{M}) &= \vec{m}^{(\ell)} \odot \vec{a}^{(\ell)}\left(\mat{W}^{(\ell)} \vec{h}^{(\ell - 1)}(\vec{x}_t; \mat{M}) + \vec{b}^{(\ell)} \right)\;, \\ %\;l=1,\ldots,L \;, \\
    p(k_t| \vec{x}_t, \mat{M}) &= 
    \frac{\exp\left({h}^{(L)}_{k_t}(\vec{x}_t; \mat{M})\right)}{\sum_j \exp\left(h^{(L)}_j(\vec{x}_t; \mat{M})\right)}\: ,
  \end{split} \label{Eq:RandomForward}
\end{equation}
where $\odot$ denotes element-wise multiplication, $\vec{a}^{(\ell)}$ is the activation function (usually a sigmoid or rectifier function) of the $\ell$-th layer, and $h^{(L)}_j$ denotes the $j$-th element in the vector function $\vec{h}^{(L)}$. The mask vectors are independent draws from a Bernoulli distribution (i.e., $p(\mat{M} | \mat{\Pi}) = \prod_{\ell, u} (\pi^{(\ell)}_u)^{m_u^{(\ell)}}(1 - \pi^{(\ell)}_u)^{1 - m_u^{(\ell)}}$) parameterized by retention probability hyperparameters, $\mat{\Pi} \defeq \{ \vec{\pi}^{(\ell)} \in [0, 1]^{D^{(\ell)}} | \ell \in \nrange{0}{L}\}$. In conventional dropout training~\cite{hinton2012improving,srivastava2014dropout}, all retention probabilities belonging to the same layer are tied (i.e., $\pi^{(\ell)}_{u} = \pi^{(\ell)}_{u'}$ for all possible $\ell$, $u$ and $u'$).

Optimizing the conditional log-likelihood of the target labels $\log p(\mathcal{K}|\mathcal{X},\Theta, \mat{\Pi})$ is intractable.
However, a tractable lower bound can be used instead;
\begin{equation}
  \begin{split}
    &\log p(\mathcal{K}|\mathcal{X},\Theta, \mat{\Pi}) \\
    &= \sum_{t} \log p(k_t | \vec{x}_t, \Theta, \mat{\Pi})  \\
    &= \sum_{t} \log \sum_{\mat{M} \in \mathcal{M}} p(k_t | \vec{x}_t, \mat{M}, \Theta) p(\mat{M}| \mat{\Pi})  \\ 
    &\geq \sum_{t} \sum_{\mat{M} \in \mathcal{M}} p(\mat{M}| \mat{\Pi}) \log p(k_t | \vec{x}_t, \mat{M}, \Theta) \;,
%\\
%    &\defeq L(\Theta; \Pi)\;,  
  \end{split} \label{Eq:Xent}
\end{equation}
where $\mathcal{M}$ is a set of all possible instances of $\mat{M}$. A straightforward application of stochastic gradient descent (SGD) applied to this lower bound leads to conventional dropout training.

%Typically, cross-entropy training of this model is performed by minimizing an upper bound $F(\Theta)$ of the cross-entropy objective function $\hat{F}(\Theta)$:
%\begin{align}
%\label{Eq:Xent}  \hat{F}(\Theta) &\defeq - \frac{1}{T} \sum_{t} \log p(k_t | \vec{x}, \Theta, \mat{\Pi})  \\
%\label{Eq:Xent-marginalization} &= - \frac{1}{T} \sum_{t} \log \sum_{\mat{M} \in \mathcal{M}} p(k_t | \vec{x}_t, \mat{M}, \Theta) p(\mat{M}| \mat{\Pi})  \\ 
%  &\leq - \frac{1}{T} \sum_{t} \sum_{\mat{M} \in \mathcal{M}} p(\mat{M}| \mat{\Pi}) \log p(k_t | \vec{x}_t, \mat{M}, \Theta) \defeq F(\Theta)\;,  \label{Eq:Xent-approximation}
%\end{align}
%where $\mathcal{M}$ is a set of all possible instances of $\mat{M}$.
%In conventional dropout training~\cite{hinton2012improving,srivastava2014dropout}, all retention probabilities belonging to the same layer are tied, i.e. $\pi^{(\ell)}_{u} = p^{(\ell)}$ for all $\ell$ and $u$, and the objective function $F(\Theta)$ is optimized with stochastic gradient descent (SGD).

\subsection{Prediction}

At test time, it is not tractable to marginalize over all mask vectors as in Eq.~\eqref{Eq:Xent}. Instead, a crude (but fast) approximation is applied. The average over all network outputs with all possible mask vectors is replaced by the output of the network with the average mask, i.e. the vector $\vec{m}^{(\ell)}$ in Eq.~\eqref{Eq:RandomForward} is replaced by the expectation vector $\vec{\pi}^{(\ell)}$:
\begin{equation}
  \begin{split}
    \tilde{\vec{h}}^{(0)}(\vec{x}_t; \mat{\Pi}) &= \vec{\pi}^{(0)} \odot \vec{x}_t, \\
    \tilde{\vec{h}}^{(\ell)}(\vec{x}_t; \mat{\Pi}) &= \vec{\pi}^{(\ell)} \odot \vec{a}^{(\ell)}\left(\mat{W}^{(\ell)} \tilde{\vec{h}}^{(\ell - 1)}(\vec{x}_t; \mat{\Pi}) + \vec{b}^{(\ell)} \right).
  \end{split}
\label{Eq:ApproxForward}
\end{equation}
In this way, we obtain an approximation of the predictive distribution $p(k|\vec{x}, \mat{\Pi}, {\Theta})$, which we denote by $\tilde{p}(k|\vec{x}, \mat{\Pi}, {\Theta})$. In practice, this approximation does not degrade prediction accuracy \cite{srivastava2014dropout}.

\section{Retention Probability Optimization} \label{Sec:ProbOpt}

For leveraging the retention probabilities as a unit pruning criterion, 
we propose to untie the retention probability parameters $\pi^{(\ell)}_{u}$ and put a bimodal prior on them.  
Then, we seek to optimize the joint log-likelihood $\log p(\mathcal{K}, \mat{\Pi} |\mathcal{X}, \Theta)$. In the next subsection, we compute the parameter gradients. In the second subsection, we describe the control variates we used to reduce variance in the gradients.  Next, we describe the prior we used to encourage the posterior retention probabilities to converge to $0$ or $1$. Finally, we summarize the algorithm.

\subsection{Stochastic Gradient Estimation}

The joint log-likelihood is given by 
%By introducing a prior distribution $p(\mat{\Pi})$ to Eq. \eqref{Eq:Xent}, an objective function for cross-entropy training of $\mat{\Pi}$ can be defined as follows:
\begin{equation}
\begin{split}
    &\log p(\mathcal{K}, \mat{\Pi} |\mathcal{X}, \Theta) \\
    &= \log p(\mathcal{K}|\mat{\Pi}, \mathcal{X}, \Theta)   + \log p(\mat{\Pi}) \\
    &= \sum_{t} \log \sum_{\mat{M} \in \mathcal{M}}  p(k_t | \vec{x}_t, \mat{M}, \Theta) p(\mat{M}| \mat{\Pi})  + \log p(\mat{\Pi}) \\
  &\defeq L(\Theta; \mat{\Pi}) + \log p(\mat{\Pi}) \;,
\end{split}
\end{equation}
where $p(\mat{\Pi})$ is the prior probability distribution of the retention probability parameters. The objective function with respect to the weight parameters ${\Theta}$ is unchanged (up to a constant), so we can follow the conventional dropout parameter updates for ${\Theta}$. %, and $L(\Pi; \Theta)$ is introduced to represent the log likelihood of $\mat{\Pi}$.
%Gradient vector of $L(\mat{\Pi}; \Theta)$ can be expressed as follows \footnote{In this paper, we interchangeably use set of parameters also as a vector that is constructed by enumerating all parameters in the parameter set.}:
The partial derivative of $L$ with respect to the retention probability of the $u$-th unit in the $\ell$-th layer is:
\begin{equation}\label{Eq:gradient-retention-probs}
  \begin{split}
    \frac{\partial L}{\partial \pi^{(\ell)}_{u}} 
    &= \sum_{t} \sum_{\mat{M} \in \mathcal{M}} p(\mat{M}| \mat{\Pi}) w_t(\mat{M}) \left( \frac{\partial }{\partial \pi^{(\ell)}_u}  \log p(\mat{M} | \mat{\Pi}) \right) \\
    &= \sum_t \E{p(\mat{M}| \mat{\Pi})}{w_t(\mat{M}) \left( \frac{\partial }{\partial \pi^{(\ell)}_u}  \log p(\mat{M} | \mat{\Pi}) \right)},
  \end{split}
\end{equation}
where the weight function $w$ is:
\begin{equation}\label{Eq:update-weights}
w_t(\mat{M}) = \frac{p(k_t | \vec{x}_t, \mat{M}, \Theta) }{\sum_{\mat{M}' \in \mathcal{M}} p(k_t | \vec{x}_t, \mat{M}', \Theta) p(\mat{M}'| \mat{\Pi})} .
\end{equation}
Similarly to prediction in the conventional dropout method, computing the denominator in the weight function is intractable due to the summation over all binary mask vectors. Therefore, we employ the same approximation, i.e. the denominator in Eq.~\eqref{Eq:update-weights} is computed using Eq.~\eqref{Eq:ApproxForward}. Hence, the weight function is approximated as:
\begin{equation}
w_t(\mat{M}) \approx \frac{p(k_t | \vec{x}_t, \mat{M}, \Theta) }{\tilde{p}(k_t | \vec{x}_t, \mat{\Pi}, {\Theta})} \defeq \tilde{w}_t(\mat{M})\;.
\label{Eq:ApproxWeight}
\end{equation}
The approximated weight function is computed by two feed forward passes: One with a stochastic binary mask and one with expectation scaling. 

The partial derivatives of $\log p(\mat{M} | \mat{\Pi})$ with respect to the retention probability parameters can be expressed as follows:
\begin{equation}
  \begin{split}
    \frac{\partial }{\partial \pi^{(\ell)}_u} \log p(\mat{M} | \mat{\Pi})
  &= \frac{m^{(\ell)}_{u}}{\pi^{(\ell)}_u} - \frac{1 - m^{(\ell)}_{u}}{1 - \pi^{(\ell)}_u}\;.
  \end{split} \label{Eq:PriorDerivative}
\end{equation}

\subsection{Variance Reduction with Control Variates}

The standard SGD approach uses an unbiased Monte Carlo estimate of the true gradient vector. However, the gradients of $L(\Theta; \mat{\Pi})$ with respect to the retention probability parameters exhibit high variance. We can reduce the estimator's variance using control variates, closely following ~\cite{ba2015learning,mnih2014neural}.
%It is applicable if a random variable, which is correlated with the true gradient and whose expectation is known, can be found.
We exploit the fact that
\begin{equation}
    \E{p(\mat{M} | \mat{\Pi})}{\frac{\partial }{\partial \pi^{(\ell)}_u} \log p(\mat{M} | \mat{\Pi}) } = 0\;,
\end{equation}
cf. Eq.~\eqref{Eq:PriorDerivative}. This implies
\begin{equation}
    \frac{\partial L}{\partial \pi^{(\ell)}_{u}} 
    = \sum_t \E{p(\mat{M}| \mat{\Pi})}{\left(w_t(\mat{M}) - C \right) \frac{\partial }{\partial \pi^{(\ell)}_u}  \log p(\mat{M} | \mat{\Pi}) } 
\end{equation}
for any $C$ that does not depend on $\mat{M}$. Thus, an unbiased estimator is given by:
\begin{equation}\label{Eq:sampled-stabilized-derivative}
  \frac{\partial L}{\partial \pi^{(\ell)}_{u}} \approx \frac{T}{|\mathcal{R}|} \sum_{r \in \mathcal{R}} \left(w_t(\mat{M}_r) - C \right) \left( \frac{\partial}{\partial \pi^{(\ell)}_u} \log p(\mat{M}_r | \mat{\Pi}) \right)\;,
\end{equation}
where $\mathcal{R}$ is a random mini-batch of training data indices and $\mat{M}_r$ is a set of mask vectors randomly drawn from $p(\mat{M} | \mat{\Pi})$ for each element in $\mathcal{R}$. As before, we approximate $w_t(\mat{M}) \approx \tilde{w}_t(\mat{M})$ to make the computation tractable.
%Denoting
%\begin{equation}
%s = w(\vec{x}, k, \mat{M}) \frac{\partial }{\partial \pi^{(\ell)}_u} \log p(\mat{M} | \mat{\Pi})
%\end{equation}
%and 
%\begin{equation}
%\delta = \frac{\partial }{\partial \pi^{(\ell)}_u} \log p(\mat{M} | \mat{\Pi})\;,
%\end{equation}
%the optimal $C$, which minimizes the variance, can be determined as
%\begin{equation}
%C = - \mathrm{Cov}[\delta, s]/\mathrm{Var}[s] \;.
%\end{equation}

The optimal $C$ does not have a closed-form solution. However, a reasonable choice for $C$ is 
\begin{equation}
    C = \E{p(\mat{M} | \mat{\Pi})}{w_t(\mat{M})} = 1 
\end{equation}
With this choice, we obtain an interpretable update rule: A training example $(x,k)$ only contributes to an update of the retention probabilities $\mat{\Pi}$ if the predictive distribution changes by applying a random dropout mask, i.e. if $p(k | \vec{x}, \mat{M}, {\Theta}) \neq \tilde{p}(k | \vec{x}, \mat{\Pi}, {\Theta})$.

\subsection{Prior distribution of retention probability}

%The remaining configuration we have to specify is a prior distribution function of retention probability $\pi^{(\ell)}_u$.
In order to encourage the posterior to place mass on compact models, a prior distribution that strongly prefers $\pi^{(\ell)}_u = 1$ or $\pi^{(\ell)}_u = 0$ is required. We use a powered beta distribution as the prior distribution
\begin{equation}
  p(\pi^{(\ell)}_u | \alpha, \beta, \gamma) \propto \left( \left(\pi^{(\ell)}_u\right)^{\alpha-1} \left(1 - \pi^{(\ell)}_u\right)^{\beta-1} \right)^{\gamma}.
\nonumber
\end{equation}
Computing the partition function is intractable when $\gamma (\alpha - 1) \leq 0$ or $\gamma (\beta - 1) \leq 0$. However, computing the partition function is not necessary for SGD-based optimization.

By setting $\alpha < 1$ and $\beta < 1$, the prior probability density goes to infinity as $\pi^{(\ell)}_u$ approaches $0$ and $1$, respectively. Thus, we can encourage the optimization result to converge to either $\pi^{(\ell)}_{u} = 0$ or $\pi^{(\ell)}_{u} = 1$. The exponent $\gamma$ is introduced in order to control the relative importance of the prior distribution. By setting $\gamma$ sufficiently large, we can ensure that the retention probabilities converge to either $0$ or $1$.

\subsection{Algorithm}

Finally, the stochastic updates of the retention probabilities with control variates are summarized in Algorithm~\ref{Algo:SGD}. 
In our experiments, we alternate optimization of the neural network parameters ${\Theta}$ and the retention probabilities $\mat{\Pi}$.
Specifically, updates computed with Algorithm \ref{Algo:SGD} are applied after each epoch of conventional dropout training.
Algorithm \ref{Algo:Alt} shows the overall structure.
%To improve generalization, we use a disjoint training set for the retention probabilities updates.
%the optimization process of the neural net parameters ${\Theta}$.
%The retention probability update algorithm computes two predictive distributions by performing two separate feed-forward processing with randomly drawn mask and without random binary mask but with expectation mask, and computes weight $\tilde{w}$ for corresponding training example by taking ratio of probabilities of the correct label obtained from these two predictive distributions.
%As described in the previous sections, the retention probabilities are updated by the gradient vector that can be expressed as the weighted sum of the gradient vectors of the logarithmic Bernoulli distributions.
%After each cross-entropy training epochs, this probability update algorithm is inserted for realizing the alternate updates of neural net parameters ${\Theta}$ and retention probabilities $\mat{\Pi}$.
After each epoch, we can remove hidden units with retention probability smaller than a threshold without degrading performance.
Therefore, we can already benefit from compaction during the training phase.

\begin{algorithm}[tb]
  \DontPrintSemicolon
  \KwData{Training data ($\mathcal{X}, \mathcal{K}$), neural net parameter ${\Theta}$, initial values $\mat{\Pi}$, random mini-batch $\mathcal{R}$}
  \KwData{Hyperparameters ($\alpha, \beta, \gamma$), learning rate $\eta$, control variate $C$}
  \KwResult{Updated dropout probabilities  $\mat{\Pi}$}
  \tcp{Derivative wrt $\log p(\mat{\Pi})$}
  \For{$\ell \in \nrange{0}{L}, u \in \nrange{1}{D^{(\ell)}}$}{
    $\delta^{(\ell)}_{u} \leftarrow {\frac{\partial}{\partial \pi^{(\ell)}_{u}} \log p(\pi^{(\ell)}_{u})}$
  }
  \tcp{Approx. derivative of $\log p(k_r | \vec{x}_r, \mat{\Pi})$}
  \For{$r \in \mathcal{R}$}{
    $\mat{M} \leftarrow $ draw a mask from $p(\mat{M}|\mat{\Pi})$ \;
    $\tilde{w} \leftarrow \frac{p(k_r| \vec{x}_r, \mat{M}, {\Theta})}{\tilde{p}(k_r| \vec{x}_r, \mat{\Pi}, {\Theta})}$ \hfill \tcp{Eq. \eqref{Eq:ApproxWeight}}\;

    $\vec{\delta} \leftarrow \vec{\delta} + (\tilde{w} - C) \vec{\nabla}_{\mat{\Pi}} \left[ \log p(\mat{M}|\mat{\Pi})\right]$ \hfill \tcp{Eq. \eqref{Eq:PriorDerivative}}\;
    %\For{$\ell \in \nrange{0}{L}, u \in \nrange{1}{D^{(\ell)}}$}{
    %  \eIf{$m^{(\ell)}_{u} = 0$} {
    %    $\delta^{(\ell)}_{u} \leftarrow \delta^{(\ell)}_{u} - (\tilde{w} + C) \frac{1}{1 - \pi^{(\ell)}_{u}}$
    %  }{
    %    $\delta^{(\ell)}_{u} \leftarrow \delta^{(\ell)}_{u} + (\tilde{w} + C) \frac{1}{\pi^{(\ell)}_{u}}$
    %  }
    %}
  }
  $\mat{\Pi} \leftarrow \mathrm{Clip}\left[\mat{\Pi} + \eta \mat{\delta}\right]$ \;
    \tcp{Clip computes $\max\{0, \min\{1, x\}\}$ for each element $x$ in the vector}
  \caption{Single update of retention probability} \label{Algo:SGD}
\end{algorithm}

\begin{algorithm}[tb]
  \DontPrintSemicolon
  \KwData{Training data ($\mathcal{X}, \mathcal{K}$), initial values ${\Theta}, \mat{\Pi}$}
  \KwResult{Updated dropout probabilities  $\mat{\Pi}$ and neural net parameters ${\Theta}$}
  \While{performance improves}{
    Optimize neural net parameters $\Theta$ (see \cite{strom2015scalable} for detail) \;
    \For{$\mathcal{R}$ in the set of random batches in $\mathcal{X}$}{
      $\mat{\Pi} \leftarrow \mathrm{Algorithm1}(\mathcal{X}, \mathcal{K}, {\Theta}, \mat{\Pi}, \mathcal{R})$
    } \;
    Remove hidden units with zero retention probability
  }
  \caption{Alternating updates of DNN parameters ${\Theta}$ and retention probabilities $\mat{\Pi}$} \label{Algo:Alt}
\end{algorithm}

\section{Experiments} \label{Sec:Experiments}

First, as a pilot study, we evaluated dropout compaction on the MNIST handwritten digit classification task. These experiments demonstrate the efficacy of our method on a widely used and publicly available dataset.
Next, we conducted a systematic evaluation of dropout compaction on three real-world speech recognition tasks.

In the experiments, we compared the proposed method against the dropout annealing method and SVD-based compaction. 
Dropout annealing was chosen because the proposed method also varies the dropout retention probabilities while optimizing the DNN weights.
The SVD-based compaction method was chosen since this technique is widely used and applicable to many tasks.

\subsection{MNIST}

\begin{table}[tb!]
  \begin{center}
\small
\caption{Classification error rates and average test-set losses on the MNIST dataset for small and large networks.} \label{Table:MNISTTable}
\begin{tabular}{l|r|rr}
	\toprule   & \#Weights & Err. rates [\%] & Avg. loss  \\
      \hline {\bf Small} & & & \\
      % 3rd left point of green, and 2nd left point of blue, 2nd (SVD)
      Baseline   & 42200   & $3.03 \pm 0.167$ & $0.156 \pm 0.014$  \\
%                 & 89400   & $2.35 \pm 0.191$ & $0.147 \pm 0.024$  \\
      Dropout    & 42200   & $3.02 \pm 0.181$ & $0.164 \pm 0.027$  \\
%                 & 89400   & $2.43 \pm 0.137$ & $0.133 \pm 0.019$  \\
      Annealing  & 42200   & $2.99 \pm 0.119$ & $0.157 \pm 0.015$  \\
%                 & 89400   & $2.25 \pm 0.171$ & $0.133 \pm 0.016$  \\
      SVD        & 40400   & $3.25 \pm 0.196$ & $0.138 \pm 0.011$  \\
                 & 82000   & $2.21 \pm 0.161$ & $0.123 \pm 0.016$  \\
      Compaction & 46665.2 & $2.55 \pm 0.188$ & $0.101 \pm 0.011$  \\
      \hline {\bf Large} & & & \\ 
      % 2nd right point of green, and 3rd right point of blue
      Baseline   & 477600   & $1.52 \pm 0.089$ & $0.0850 \pm 0.004$ \\
%                 & 1275200  & $1.47 \pm 0.046$ & $0.0760 \pm 0.004$ \\
      Dropout    & 477600   & $1.50 \pm 0.084$ & $0.0818 \pm 0.006$ \\
%                 & 1275200  & $1.48 \pm 0.072$ & $0.0758 \pm 0.005$ \\
      Annealing  & 477600   & $1.60 \pm 0.091$ & $0.0900 \pm 0.005$ \\
%                 & 1275200  & $1.48 \pm 0.064$ & $0.0825 \pm 0.004$ \\
      SVD        & 357600   & $1.54 \pm 0.083$ & $0.0858 \pm 0.004$ \\
                 & 795200   & $1.50 \pm 0.030$ & $0.0771 \pm 0.004$ \\
      Compaction & 481276.7 & $1.49 \pm 0.039$ & $0.0536 \pm 0.003$ \\
      \bottomrule
    \end{tabular}    
  \end{center}
\end{table}

The MNIST dataset consists of 60\,000 training and 10\,000 test images ($28{\times}28$ gray-scale pixels) of handwritten digits. For simplicity, we focus on the permutation invariant version of the task without data augmentation.

We used 2 layer neural networks with rectified linear units (ReLUs). The parameters of the DNNs were initialized with random values from a uniform distribution with adaptive width computed with Glorot's formula ~\cite{glorot2010understanding} \footnote{We also evaluated with the ReL variant of Glorot's initialization~ \cite{he2015delving}; however, the ReL variant did not outperform the original Glorot initialization in our experiments.}.

As is standard, we split the training data into 50\,000 images for training and 10\,000 images for hyperparameter optimization.
Learning rates, momentum, and the prior parameters were selected based on development set accuracy. The mini-batch size for stochastic gradient descent was set to $128$.
We evaluated networks with various numbers of the hidden units $D^{(\ell)} \in \{25, 50, 100, 200, 400, 800, 1600\}$.
For dropout compaction training, we set $\beta = \alpha$, to produce approximately 50\% compression.
The SVD compacted models were trained by applying SVD to the hidden-to-hidden weight matrices in the best performing neural networks in each configuration of $D^{(\ell)}$. The sizes of the bottleneck layer were set to $\lceil D^{(\ell)} / 8 \rceil$ to achieve 25\% compression in terms of the number of the parameters in the hidden-to-hidden matrix. After the decomposition, the SVD-compacted networks were again fine-tuned to compensate for the approximation error.

Based on manual tuning over the the development set accuracies, the learning rate and momentum were set to $0.001$ and $0.9$ respectively. L2 regularization was found not to be effective for the baseline system. The optimal L2 regularization constants were $10^{-6}$ for dropout and annealed dropout and $10^{-4}$ for dropout compaction.
For the dropout annealing method, we increased the retention probability from $0.5$ to $1.0$ over the first 4 epochs.

\begin{figure*}[bt!]
\begin{center}
\includegraphics[height=3.0in]{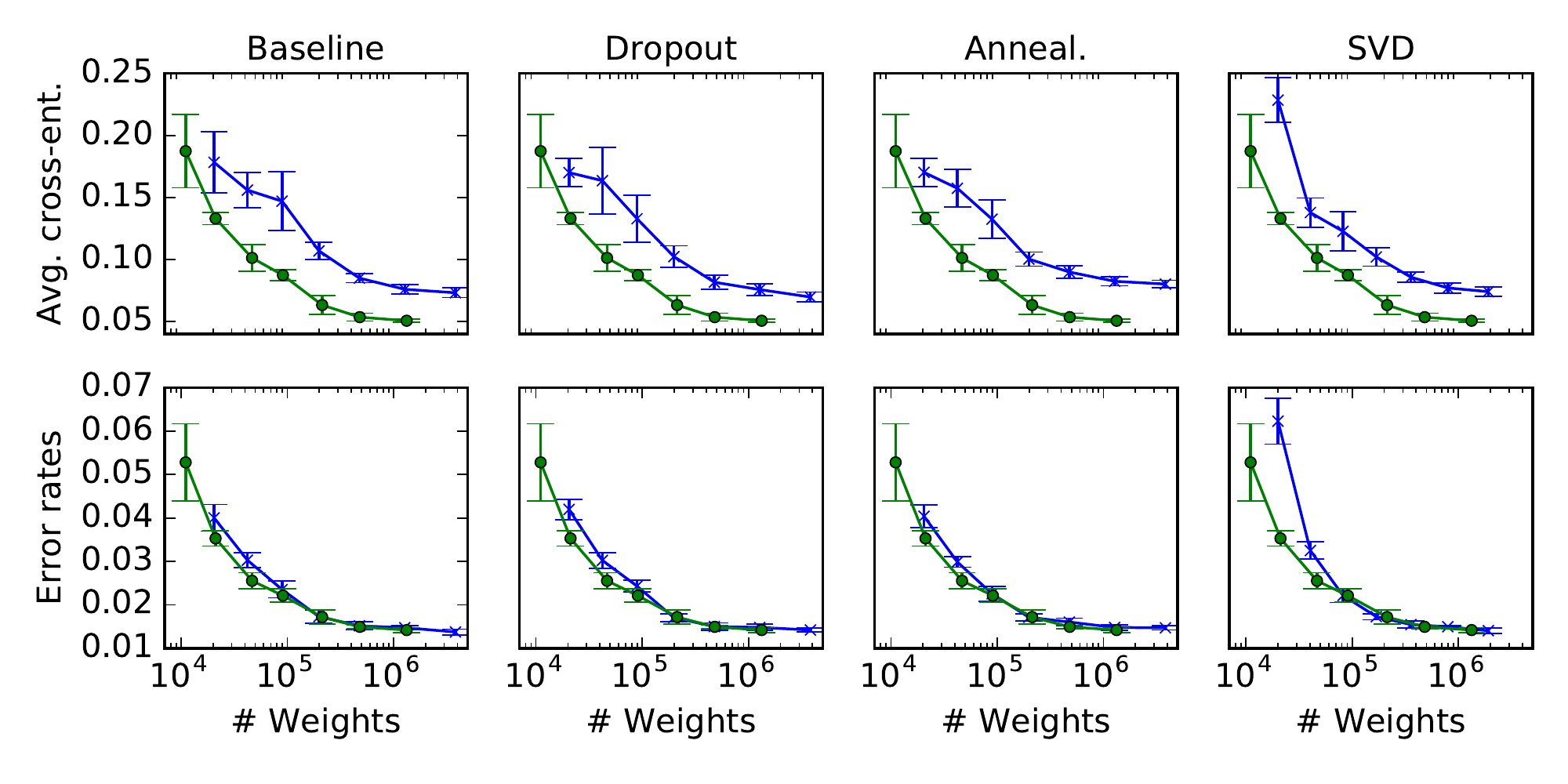} 
\caption{The averages test-set loss (top) the number of test-set errors (bottom) in the MNIST experiments as functions of the numbers of weights. Each column shows the results of dropout compaction compared to a conventional feed-forward network, conventional dropout, dropout annealing, and singular value decomposition, respectively. The green lines with {\bf o} marker and blue lines with {\bf x} marker denote the results of proposed and the compared method, respectively. The error bars show $\pm 1 \sigma$.} \label{Fig:MNIST}
\end{center}
\end{figure*}

Figure \ref{Fig:MNIST} and Table \ref{Table:MNISTTable} show the results of our proposed method and the other methods in comparison.
In the figure, the plots in the first row show the differences in the average cross-entropy loss computed on the test set, and the plots in the second row show the differences in the classification error rate. The green lines with ``{\bf o}'' markers denote the proposed method and the blue lines with ``{\bf x}'' markers denote the compared method, i.e. baseline feed forward net, conventional dropout, dropout annealing, and SVD compacted DNNs. The error bars in the plots represent two standard deviations ($\pm 1 \sigma$) estimated from 10 trials  with different random initialization.
The table shows the results for small and large networks. The numbers of the weights in the table are the average numbers of the trials.

In terms of test-set cross-entropy loss, dropout compaction performs consistently better than the other methods in comparison. 
For the application to automatic speech recognition, which is our main interest, the performance in terms of cross-entropy loss is decisive, because the DNN is used as an estimator for the label probability (rather than a classifier).

The behavior in terms of error rate differs. 
By increasing the model size, the error rate eventually saturates at the same point for all methods. However, with small networks, dropout compaction also clearly outperforms the other methods in terms of error rate. The case of small networks is the more relevant one, because our aim is to apply dropout compaction for training small models. Here, "small" must be understood relative to the complexity of the task. Neural networks, which can be deployed in large-scale production for difficult tasks like speech recognition, can typically be considered as small.

%We observed that dropout compaction outperformed the other method in most cases even though the error rates were already saturated especially if the sufficiently large networks were used.
%However, for the smaller networks, it was clearly shown that the proposed method performed better than the other methods in comparison.
%This property is an advantage in complicated tasks since, the relative complexity of networks will be smaller in such complicated problems.
%The relative advantage of the proposed method in average cross-entropy losses was clearly shown.
%Since our main interest is automatic speech recognition where DNNs are used as an estimator of the label distribution, improvement in cross-entropy is important since it implies better estimation of the target distribution.

%To deploy neural network predictions at a scale, it is important to keep the networks as small as possible; therefore, the relative advantage of these methods is expected to be larger in the real-world experiments than that in the MNIST experiments.

\subsection{Large-Vocabulary Continuous Speech Recognition}

\begin{table*}[tb!]
\begin{center}
\small
\caption{Word error rates [\%] on evaluation sets of the speech recognition tasks; the development set error rates are shown in parentheses.} \label{Table:WERs}
\begin{tabular}{ll|rr|rr|rr}
  \toprule
  & & Baseline & Annealing & SVD & Compaction  & SVD & Compaction \\
  \hline
  & \# Hid. unit & 1536 & 1536 & (3072, 384) & 3072 $\to$  \textasciitilde 1536 & (1536, 192) & 1536 $\to$  \textasciitilde 768 \\
  \hline
  {VoiceSearchSmall} & XEnt & 22.5 (22.6) & 22.3 (21.7) & 22.5 (22.3) & {\bf 21.8} ({\bf 21.2}) & 22.7 (21.6) & 22.6 (21.7)\\
  & + bMMI                  & 21.2 (21.0) & 20.0 (19.8) & 21.5 (21.1) & {\bf 19.9} ({\bf 19.7}) & 20.8 (20.3) & 20.4 (19.8)\\
  {VoiceSearchLarge} & XEnt & 18.9 (18.2) & 18.7 (18.2) & 18.8 (18.2) & {\bf 18.4} ({\bf 18.0}) & 19.2 (18.8) & 18.9 (18.5) \\
  & + bMMI                  & 17.9 (17.4)  & 17.2 ({\bf 17.0}) & 17.7 (17.2) & {\bf 16.9} ({\bf 17.0}) & 17.3 (17.2) & 17.3 (17.2)\\
  {GenericFarField} & XEnt  & 24.0 (21.9) & 23.6 (21.7) &  24.4 (22.5) & {\bf 23.4} ({\bf 21.3}) & 24.8 (22.8) & 23.9 (22.1) \\
   & + bMMI                 & 21.4 (19.7) & {\bf 20.6} (18.8) & 21.3 (19.1) & 20.8 ({\bf 18.7}) & 20.8 (18.9) & 21.0 (18.9)  \\
  \bottomrule
\end{tabular}
\end{center}
\end{table*}

\begin{figure}[bt!]
\begin{center}
\includegraphics[height=1.5in]{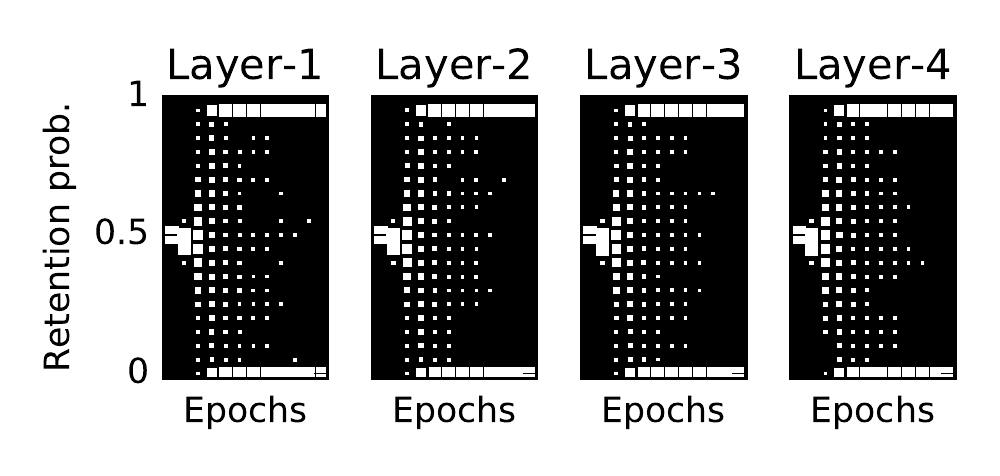}
\end{center}
\caption{Evolution of histograms over retention probability in the first 12 epochs on VoiceSearchSmall.} \label{Fig:Hists}
\end{figure}

\begin{figure}[bt!]
\begin{center}
\includegraphics[height=1.9in]{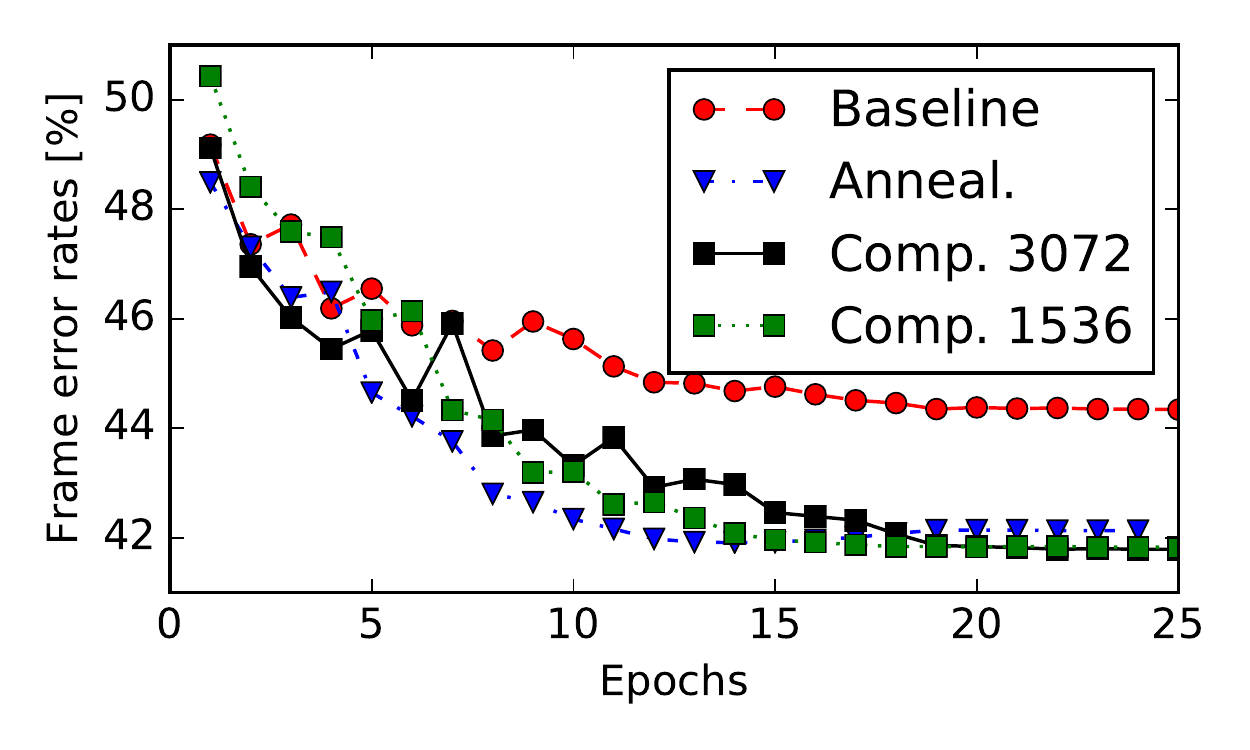}
\end{center}
\caption{Frame error rates as a function of the number of epochs on VoiceSearchSmall.} \label{Fig:FERs}
\end{figure}

As an example of a real-world application, which requires large-scale deployment of neural networks,
we applied dropout compaction to large-vocabulary continuous speech recognition (LVCSR). We performed experiments on three tasks:
VoiceSearchLarge, which contains 453h of voice queries, and VoiceSearchSmall, which is a 46h subset of VoiceSearchLarge. GenericFarField contains 115h of far-field speech, where the signal is obtained by applying front-end processing to the seven channels from from a microphone array.
We used VoiceSearchSmall for conducting preliminary experiments for finding the optimal hyperparameters, and used these for the other tasks.

As input vectors to the DNNs, we extracted 32 dimensional log Mel-filterbank energies over 25ms frames every 10ms. 
The DNN acoustic model processed 8 preceding, a middle frame, and 8 following frames as a stacked vector (i.e., $32 \times 17 = 544$ dimensional input features for each target). Thus, with our feature extraction, the number of training examples is 16.5M (VoiceSearchSmall), 163M (VoiceSearchLarge), and 41M (GenericFarField),  respectively.

We used a random 10\% of the training data as a validation set, which was used for ``Newbob''-performance based learning rate control \cite{QuickNet}. Specifically, we halved the learning rate when the improvement from the last epoch is less than a threshold.
As an analogy of cross validation-based model selection, we used 10\% of the validation set for optimizing the retention probabilities.
%Therefore, the frame error rates shown below may not be suitable as a measure of generalization ability because those are computed on the validation set. However, the development and evaluation sets are disjoint from the validation set.
%Thus, the word error rates computed on the development and evaluation sets should reflect the generalization ability of the proposed method.

The baseline model size was designed such that the total ASR latency was below a certain threshold.
The number of hidden units for each layer was determined to be 1\,536 and the number of hidden layers was 4. The sigmoid activation function was used for nonlinearity in the hidden layers.
Following the standard DNN/HMM-hybrid modeling approach, the output targets were clustered HMM-states obtained via triphone clustering. We used 2\,500, 2\,506, and 2\,464 clustered states with the VoiceSearchSmall, VoiceSearchLarge, and GenericFarField tasks, respectively.
For fast evaluation, we quantized the values in the weight matrices and used integer operations for the feed-forward computations.
These networks are sufficiently small for achieving low latency in a speech recognition service.
Therefore, in the experiments, we focus on two use cases: (a) Enabling the use of a larger network within the given fixed budget, and (b) achieving faster evaluation by compressing the current network.

All networks were trained with distributed parallel asynchronous SGD on 8 GPUs~\cite{strom2015scalable}. In addition, all networks were pretrained with the greedy layer-wise training method~\cite{bengio2007greedy}.
For the dropout compaction and annealing methods, retention probabilities were kept fixed to 0.5 during pretraining.
For annealed dropout, we use a schedule designed to increase the retention probabilities from 0.5 to 1.0 in the first 18 epochs.
As in the MNIST experiments, we fixed the hyperparameters to obtain approximately 50\% compression
\footnote{50\% compression of hidden units yields roughly 25\% compression of hidden-to-hidden weight matrices and 50\% compression of the input and output layer weight matrices. This leads to a roughly 2.5x speedup in the feed-forward computation.} 
by setting $\alpha = \beta = 0.9$, and $\gamma = T'$ where $T'$ is the number of non-silence frames in the data set.

Because conventional dropout did not improve the performance of the baseline system in the ASR experiments, we only use annealed dropout as the reference system for dropout-based training.
This may be due to the small size of the neural networks relative to our training corpus size.

\begin{comment}
\begin{table*}[tb!]
\begin{center}
\small
\caption{Frame error rates over the cross-validation sets} \label{Table:FERs}
\begin{tabular}{r|rrrr}
  \toprule
  & Baseline & Annealing & Compaction  & Compaction \\
  \hline
  \# hid. unit& 1536 & 1536 & 3072 $\to$ \textasciitilde 1536 & 1536 $\to$  \textasciitilde 768 \\
  \hline
  VoiceSearchSmall & 44.3 & 41.9 & {\bf 41.8} & {\bf 41.8} \\
  VoiceSearchLarge & 40.9 & 40.8 & {\bf 39.8} & 40.7 \\
  GenericFarField & 56.4 & {\bf 53.7} &  54.5  & 54.6 \\
  \bottomrule
\end{tabular}
\end{center}
\end{table*}
\end{comment}

% GenericFarField
% input: 17 * 32 = 544
% output: 2464
% H=1536: (544 + 1) * 1536 + (1536 + 1) * 1536 * 3 + (1536 + 1) * 2464 = 11706784
% Large: (544 + 1) * 1580 + (1580 + 1) * 1497 + (1497 + 1) * 1539 + (1539 + 1) * 1500 + (1500 + 1) * 2464
% Small: (544 + 1) * 759 + (759 + 1) * 746 + (746 + 1) * 766 + (766 + 1) * 737 + (737 + 1) * 2464

%Table \ref{Table:FERs} shows the best frame error rates (FERs) we achieved for the validation set.
%We observed that the proposed method did not degrade the FERs even when we compressed the number of hidden units from 1536 to roughly 50\%.
%In the VoiceSearch experiments, FER results were even better than using annealed dropout even when the number of hidden units is reduced.
%Furthermore, we observed that starting from a 200\% larger model yielded better performance than training the same size DNN with or without annealed dropout.
%As expected, larger models performed better on the larger corpus.

Fig.~\ref{Fig:Hists} shows histograms of the retention probabilities as functions of the numbers of epochs on the VoiceSearchSmall task.
As designed, the retention probabilities (initialized to 0.5) rapidly diffused from the initial value $0.5$, and converged to $0$ or $1$ in the first 11 epochs.
We did not observe significant differences with regard to the pruning rate and the convergence speed over different hidden layers. The compression rates of hidden layers were around 50\% in all layers.

Fig.~\ref{Fig:FERs} shows the evolution of the frame error rate on the VoiceSearchSmall task.
We observed that annealed dropout started overfitting in the later epochs, after the retention probability was annealed to $1$.
On the other hand, the dropout compaction methods exhibited the performance gain after the probabilities were converged completely.
This suggests that, similar to SVD-based compaction, our proposed method requires some fine-tuning after the structure is fixed, even though the fine-tune and compaction processes are smoothly connected in the proposed method. 
This might be the reason why the optimal prior parameters for dropout compaction, which were selected on the development set, implied that the retention probabilities converge already in only 11 epochs, whereas the optimal parameters for dropout annealing yield a deterministic model after 18 epochs.

%This might be a reason why the optimal configuration for the dropout compaction technique implied a faster annealing process than that of dropout annealing technique.
%From the histograms, we observed that the outcome of compaction was strongly controlled by the prior hyperparameter settings. The overall compression rate was also around 50\%.
%This suggests that the prior distribution we used might be too steep. 
%However, the prior parameters were optimized by the development set, i.e. the use of smoother prior did not contribute to the development set performance improvements.
%We leave investigation of this unintuitive behavior to future work.

Table~\ref{Table:WERs} shows the word error rates over the development and evaluation sets.
As is standard in ASR, all cross-entropy models were fine-tuned according to a sequence-discriminative criterion, in this case the boosted maximum mutual information (bMMI) criterion~\cite{povey2008boosted,kingsbury2009lattice,vesely2013sequence}.
Dropout compaction has been applied in the cross-entropy phase of the optimization, and the pruned structure was then used in the bMMI training phase.

The results in Table~\ref{Table:WERs} show that dropout compaction models starting with a larger structure (use case (a)) yielded the best error rates in all cases except for the bMMI result on the GenericFarField task and they always performed better than the baseline.
The differences were especially large in comparison to the cross-entropy trained models.
The reason for this might be that the dropout compaction method determines the structure based on the cross-entropy-based criterion.

Regarding use case (b), dropout compaction achieved better results than the baseline network on all tasks. Further, most of the gains by annealed dropout are retained, although the dropout compaction models are roughly 2.5 times smaller.
Compared to SVD-based model compaction, the proposed method performed better in almost all cases.
Similar to the comparison in use case (a), the relative advantage of dropout compaction became smaller with the additional bMMI training step.
Therefore, adapting the proposed method to be compatible with sequence discriminative training such as bMMI is a promising future research direction.

\section{Conclusion} \label{Sec:Conclusion}

In this paper, we introduced dropout compaction, a novel method for training neural networks, which converge to a smaller network starting from a larger network. At At the same time, the method retains most of the performance gains of dropout regularization.

The method is based on the estimation of unit-wise dropout probabilities with a sparsity-inducing prior. On real-world speech recognition tasks, we demonstrated that dropout compaction provides comparable accuracy even when the final network has fewer than 40\% of the original parameters. Since computational costs scale proportionally to the number of parameters in the neural network, this results in a 2.5x speed up in evaluation.

% THIS IS THE CHANGED PARAGRAPH, NOT SURE , YOU NEED TO DECIDE WHAT YOU WANT TO MAKE OUT OF IT .. 
% DON'T MAKE IT TOO COMPLICATED, OTHERWISE NOBODY WILL UNDERSTAND IT ... ;-)
In future work, we want to study whether the results by dropout compaction can be further improved by using more sophisticated methods for estimating the expectation over the mask patterns. Further, the application of our proposed method to convolutional and recurrent neural networks is a promising direction.

\footnotesize
\bibliographystyle{IEEEtran}
\bibliography{Compaction}

\end{document}